\documentclass[3p,,preprint,12pt]{elsarticle}
\makeatletter\if@twocolumn\PassOptionsToPackage{switch}{lineno}\else\fi\makeatother

\usepackage{hyperref}  
\hypersetup{
    colorlinks=true,
    linkcolor=blue,
    urlcolor=blue,
    }
    
\usepackage{tabulary,xcolor}
\usepackage{amsfonts,amsmath,amssymb}
\usepackage[T1]{fontenc}
\usepackage{placeins}
\usepackage{soul}

\makeatletter
\let\save@ps@pprintTitle\ps@pprintTitle
\def\ps@pprintTitle{\save@ps@pprintTitle\gdef\@oddfoot{\footnotesize\itshape \null\hfill\today}}
\def\hlinewd#1{%
  \noalign{\ifnum0=`}\fi\hrule \@height #1%
  \futurelet\reserved@a\@xhline}

\AtBeginDocument{\ifNAT@numbers \biboptions{sort&compress}\fi}
\makeatother

\usepackage{ifluatex}
\ifluatex
\usepackage{fontspec}
\defaultfontfeatures{Ligatures=TeX}
\usepackage[]{unicode-math}
\unimathsetup{math-style=TeX}
\else 
\usepackage[utf8]{inputenc}
\fi 
\ifluatex\else\usepackage{stmaryrd}\fi

\usepackage{url,multirow,morefloats,floatflt,cancel,tfrupee}
\makeatletter

\AtBeginDocument{\@ifpackageloaded{textcomp}{}{\usepackage{textcomp}}}
\makeatother
\usepackage{colortbl}
\usepackage{xcolor}
\usepackage{pifont}
\usepackage[nointegrals]{wasysym}
\makeatletter

\def\mcWidth#1{\csname TY@F#1\endcsname+\tabcolsep}

\def\cAlignHack{\rightskip\@flushglue\leftskip\@flushglue\parindent\z@\parfillskip\z@skip}
\def\rAlignHack{\rightskip\z@skip\leftskip\@flushglue \parindent\z@\parfillskip\z@skip}

\@ifundefined{etal}{}{}

\usepackage{ifxetex}
\ifxetex\else\if@twocolumn\@ifpackageloaded{stfloats}{}{\usepackage{dblfloatfix}}\fi\fi

\AtBeginDocument{
\expandafter\ifx\csname eqalign\endcsname\relax
\def\eqalign#1{\null\vcenter{\def\\{\cr}\openup\jot\m@th
  \ialign{\strut$\displaystyle{##}$\hfil&$\displaystyle{{}##}$\hfil
      \crcr#1\crcr}}\,}
\fi
}

\AtBeginDocument{%
  \@ifpackageloaded{endfloat}%
   {\renewcommand\efloat@iwrite[1]{\immediate\expandafter\protected@write\csname efloat@post#1\endcsname{}}}{\newif\ifefloat@tables}%
}%

\def\BreakURLText#1{\@tfor\brk@tempa:=#1\do{\brk@tempa\hskip0pt}}
\let\lt=<
\let\gt=>
\def\processVert{\ifmmode|\else\textbar\fi}

\@ifundefined{subparagraph}{
\def\subparagraph{\@startsection{paragraph}{5}{2\parindent}{0ex plus 0.1ex minus 0.1ex}%
{0ex}{\normalfont\small\itshape}}%
}{}

\newcommand\role[1]{\unskip}
\newcommand\aucollab[1]{\unskip}
  
\@ifundefined{tsGraphicsScaleX}{\gdef\tsGraphicsScaleX{1}}{}
\@ifundefined{tsGraphicsScaleY}{\gdef\tsGraphicsScaleY{.9}}{}
\def\checkGraphicsWidth{\ifdim\Gin@nat@width>\linewidth
	\tsGraphicsScaleX\linewidth\else\Gin@nat@width\fi}

\def\checkGraphicsHeight{\ifdim\Gin@nat@height>.9\textheight
	\tsGraphicsScaleY\textheight\else\Gin@nat@height\fi}

\def\fixFloatSize#1{}
\let\ts@includegraphics\includegraphics

\def\inlinegraphic[#1]#2{{\edef\@tempa{#1}\edef\baseline@shift{\ifx\@tempa\@empty0\else#1\fi}\edef\tempZ{\the\numexpr(\numexpr(\baseline@shift*\f@size/100))}\protect\raisebox{\tempZ pt}{\ts@includegraphics{#2}}}}

\AtBeginDocument{\def\includegraphics{\@ifnextchar[{\ts@includegraphics}{\ts@includegraphics[width=\checkGraphicsWidth,height=\checkGraphicsHeight,keepaspectratio]}}}

\DeclareMathAlphabet{\mathpzc}{OT1}{pzc}{m}{it}

\def\URL#1#2{\@ifundefined{href}{#2}{\href{#1}{#2}}}

\def\UrlOrds{\do\*\do\-\do\~\do\'\do\"\do\-}%
\g@addto@macro{\UrlBreaks}{\UrlOrds}

\edef\fntEncoding{\f@encoding}

\makeatother

\newif\ifmultipleabstract\multipleabstractfalse%
    \makeatletter
\def\ead{\@ifnextchar[{\@uad}{\@ead}}
\gdef\@ead#1{\bgroup
   \def\_{\string\underscorechar\space}
   \def\{{\string\lbracechar\space}
   \def\textdagger{\string\textdagger\space}
   \def\texttildeapprox{\string\texttildeapprox\space}
   \def~{\hashchar\space}
   \def\}{\string\rbracechar\space}
   \edef\tmp{\the\@eadauthor}
   \immediate\write\@auxout{\string\emailauthor
     {#1}{\expandafter\strip@prefix\meaning\tmp}}
  \egroup
}
\gdef\emailauthor#1#2{\stepcounter{ead}
      \g@addto@macro\@elseads{\raggedright
      \let\corref\@gobble
      \eadsep\texttt{#1} (#2)
      \def\eadsep{\unskip,\space}}
}

\makeatother
  
\begin{document}

\begin{frontmatter}

    \title{
  Bayesian Networks and Machine Learning for COVID-19 Severity Explanation and Demographic Symptom Classification    
}
\author{Oluwaseun T. Ajayi\corref{c-920b5d4215fc}}\cortext[c-920b5d4215fc]{Corresponding author.}
\author{Yu Cheng}
    
\address{Department of Electrical and Computer Engineering\unskip, 
    Illinois Institute of Technology\unskip, Chicago\unskip, 60616\unskip, Illinois\unskip, USA}
  	

\begin{abstract}
With the prevailing efforts to combat the coronavirus disease 2019 (COVID-19) pandemic, there are still uncertainties that are yet to be discovered about its spread, future impact, and resurgence. In this paper, we present a three-stage data-driven approach to distill the hidden information about COVID-19. The first stage employs a Bayesian network structure learning method to identify the causal relationships among COVID-19 symptoms and their intrinsic demographic variables. As a second stage, the output from the Bayesian network structure learning, serves as a useful guide to train an unsupervised machine learning (ML) algorithm that uncovers the similarities in patients' symptoms through clustering. The final stage then leverages the labels obtained from clustering to train a demographic symptom identification (DSID) model which predicts a patient's symptom class and the corresponding demographic probability distribution. We applied our method on the COVID-19 dataset obtained from the Centers for Disease Control and Prevention (CDC) in the United States. Results from the experiments show a testing accuracy of 99.99\%, as against the 41.15\% accuracy of a heuristic ML method. This strongly reveals the viability of our Bayesian network and ML approach in understanding the relationship between the virus symptoms, and providing insights on patients' stratification towards reducing the severity of the virus. Code is available at \href{https://github.com/Seunaj/Covid-19-Bayesian-Networks-CPDs}{https://github.com/Seunaj/Covid-19-Bayesian-Networks-CPDs}.

\end{abstract}
    
  \end{frontmatter}

\section{Introduction}
\label{sec:introduction}
The unexpected havoc caused by the coronavirus disease 2019 (COVID-19) has made the society and government to prioritize health and wellness than ever before. 
The current COVID-19 pandemic which was discovered in Wuhan China in 2019 \cite{adetiba2022deepcovid}, is caused by the severe acute respiratory syndrome coronavirus 2 (SARS-CoV-2) virus, and since its emergence, lives have been lost, and the economy of nations has been greatly affected. 
Sadly, the wide spread of the disease has influenced the social well-being of people; this is because the virus spreads between people who are in proximity, especially in poor ventilated rooms \cite{saadat2020environmental}. 
People often get infected when they touch their eyes, nose, or mouth after contact with an infected surface. We can therefore assert that it is difficult for individuals to take strict actions to protect themselves against the virus, and even more challenging for governments to accurately monitor, predict (or forecast) and prevent the spread.

No doubt, efforts have been made to curb the spread of the virus; such efforts include providing safety guidelines on social distancing, use of face mask, and frequent hand washing. Most importantly, vaccines have proven to be effective in alleviating the increasing spread of the COVID-19 disease.
Considering the COVID-19 mitigation strategies employed by the United States (US) government, there is need to critically study the virus cases, especially the health symptoms associated with them and the severity to respiratory failure or death \cite{amon2020covid}.
According to the US centers for disease control and prevention (CDC) \cite{covidtracker}, as of December 11, 2023, there are over 103.43 million confirmed cases and 1.10 million deaths caused by COVID-19 in the US. Notably, some states have hot spots for the virus.
The statistical data on the global, continental, national, state, and county spread of the virus is available on the CDC repository and website \cite{covidtracker}.

The severity of the impact of COVID-19 on the health and social behavior of people has recently continued to increase, even though between 2021 and 2022, there was a reduction in the spread of the initial variants through the administering of vaccines \cite{covidvaccine} - Pfizer-BioNTech, Moderna, Novavax, and Johnson \& Johnson’s, as well as contact tracing.
The recent increase in the spread of the virus stems from the emergence of new variants, some of which pose greater threats than the early ones; this is because it takes some time to study the variants, its symptoms, diagnosis, how the human respiratory system reacts to it, and how to develop a potent vaccine against it. 
Apparently, the new variants have caused health agencies (e.g., CDC, NIH and WHO) to advocate for the booster version of the available vaccines; this requires that people get multiple doses of the COVID-19 vaccines to enjoy better health in the face of the growing spread of the new variants.

It is possible that if the society has an understanding about the COVID-19 disease, its symptoms, how it spreads, and the degree of its severity across different demographics, they can better plan their lives and become resilient to the threat it poses. 
However, such public awareness relies on availability and access to historical data, and knowledge about the disease. 
In most cases, only the health agencies and government are privy to the data, while the society is served with “second-hand” information about the projection (forecasting), and safety measures for curtailing the disease. 
Since there are several symptoms associated with COVID-19, it is important to study the relationship between the symptoms and the disease. Such information can guide the government, health agencies and the society in policy making, vaccine development and responsiveness to safety, respectively. 

This paper seeks to address the information and knowledge awareness limitation about the COVID-19 disease. Other than the ``inaccessibility to data'' constraint, methods of extracting useful knowledge that is of public interest have not been studied holistically. To address these critical public and government concerns about COVID-19, we propose a data-driven method by: 
\begin{itemize}
    \item developing a probabilistic graphical model from COVID-19 dataset, where the nodes represent symptoms (i.e., variables or features), and the edges represent the causal relationship between them. The idea is that, since there are several symptoms collected for patients’ cases, it is intuitive to study the (in)dependence mapping between these features;
    \item developing an unsupervised machine learning (ML) algorithm that learns (and extracts) the similarities between the patients' symptoms and groups them into appropriate classes (or clusters). The task of the ML algorithm is unsupervised as it requires no human intervention on labeling the patients’ cases according to some prior knowledge. The idea is to learn unknown patterns in patients' symptoms which might reflect their demographics; and
    \item developing a supervised learning algorithm which can accurately predict a patient's demographic symptom class (i.e., age and gender) given the patient's COVID-19 symptoms as input.
\end{itemize}

\begin{figure*}
\centering
  \includegraphics[width=0.8\textwidth]{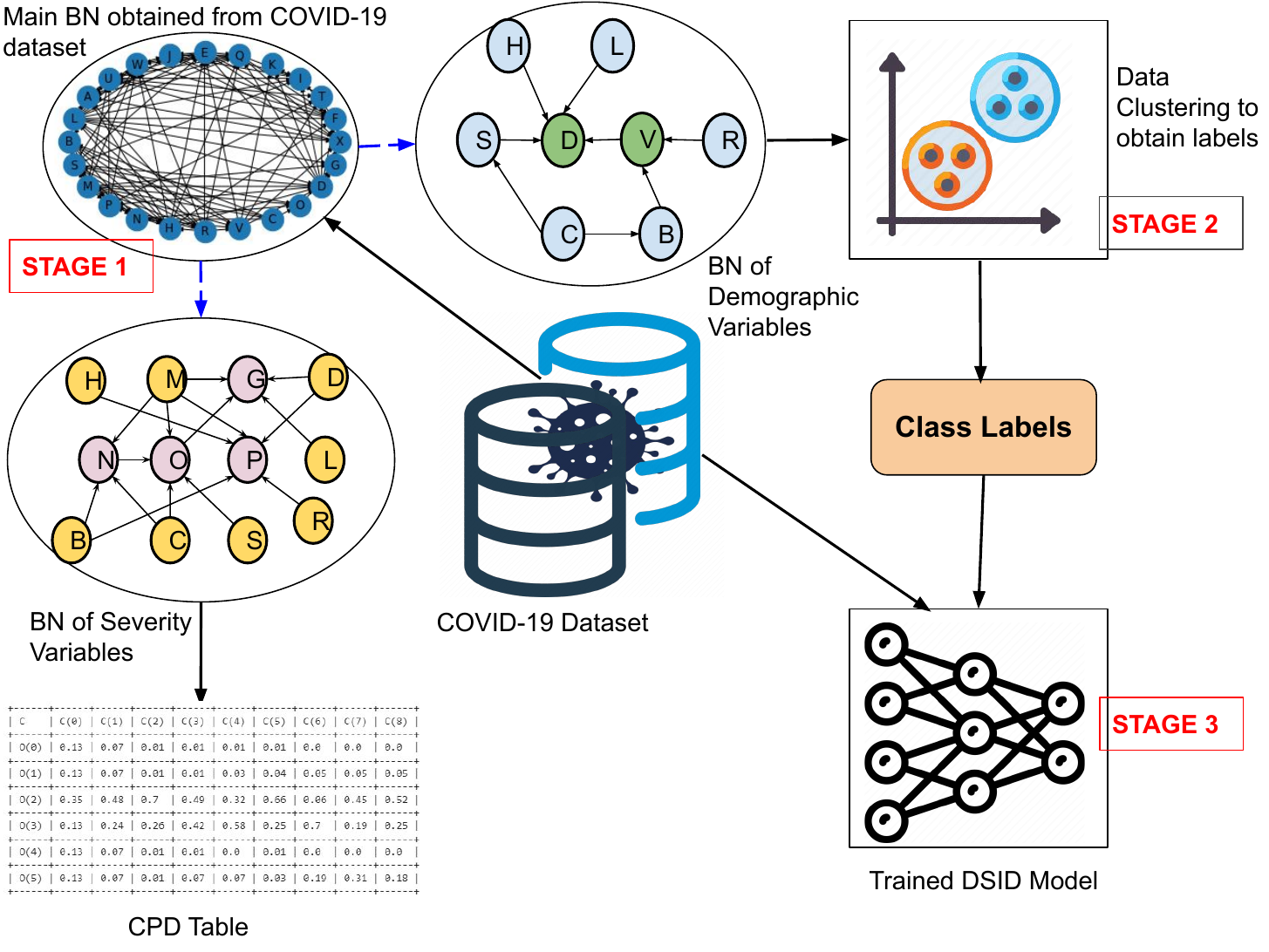}
  \caption{Proposed three-stage framework of BN and ML for COVID-19 severity explanation and demographic classification. In stage 1, the BNs of the severity variables and demographic variables are obtained from the main BN (indicated by the blue broken lines connecting them).}\label{fig:main}
\end{figure*}

Our proposed three-stage data-driven framework shown in Figure \ref{fig:main} provides a step-by-step procedure to: STAGE 1 - learn the structure of the COVID-19 dataset based on the conditional dependencies of the symptoms, and learning the symptoms' probability distributions over their dependencies; STAGE 2 - select relevant symptoms from the learned dependencies of the demographic variables to identify similarities using clustering, and STAGE 3 - training an identification model that predicts the class of patients' symptoms. This framework offers some contributions in the following ways:
\begin{itemize}
    \item vaccines can be improved through knowledge from the causal relationships among symptoms in the COVID-19 dataset. For example, if certain symptoms have a high probability of causing COVID-19 deaths and/or other diseases, such symptoms can be further studied to understand their pathology;
    \item knowledge from clustering the patients’ symptoms can help to understand patients' stratification with respect to the treatment of the disease.
\end{itemize}

\begin{figure*}
\centering
  \includegraphics[width=0.8\textwidth]{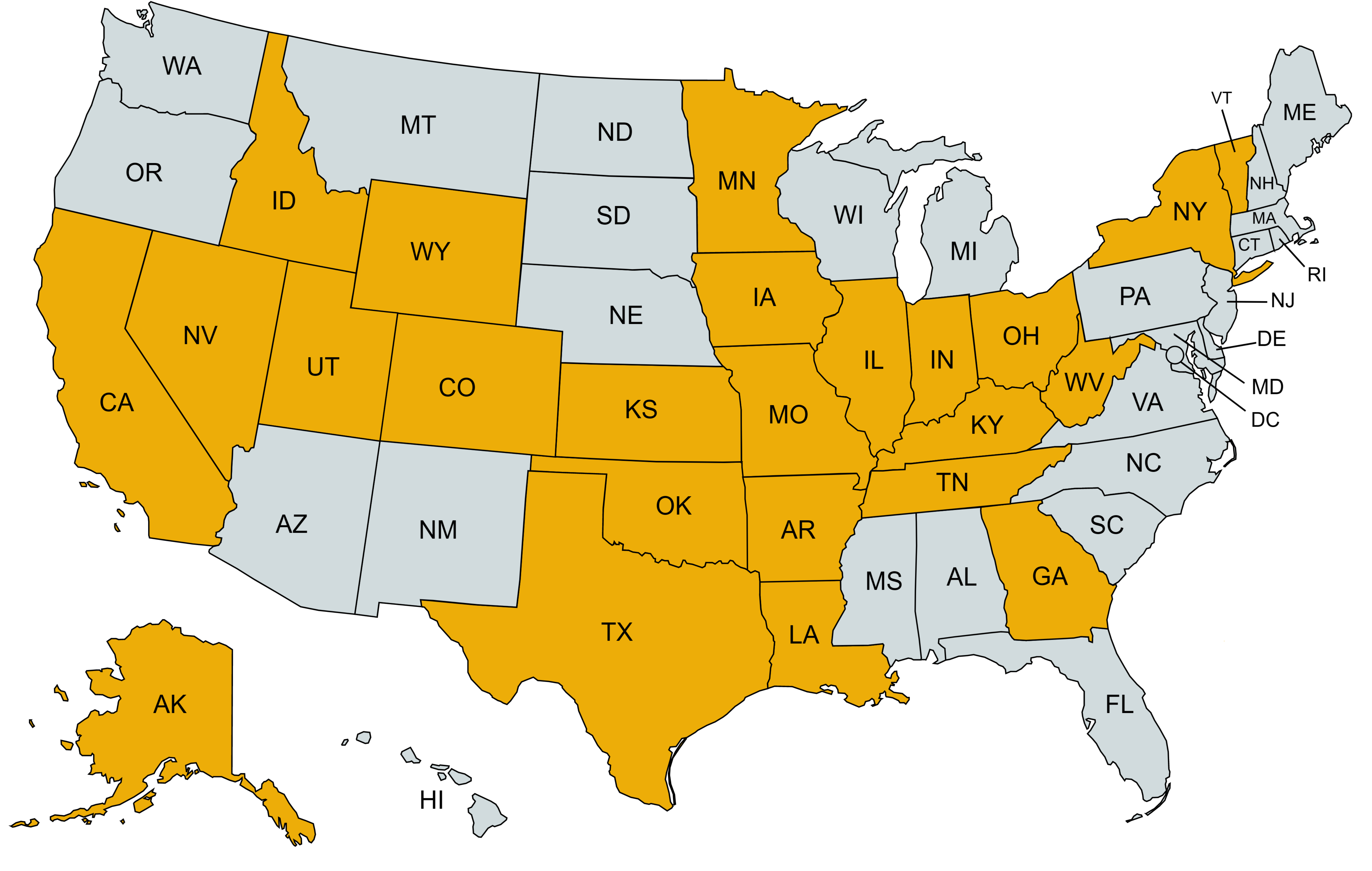}
  \caption{Data collection map for cases contained in the COVID-19 dataset provided by CDC. States with cases in the dataset are the orange ones, while those that are grey have no cases in the dataset.}\label{fig:dataset}
\end{figure*}

In this paper, we collected a total of 537,243 real patients' COVID-19 data from the US CDC\footnote[1]{The CDC does not take responsibility for the scientific validity or accuracy of methodology, results, statistical analysis, or conclusions presented. The agency only provided the data for research purposes.}; the earliest case in our dataset was recorded on January 2020 and the most recent one was recorded on October 2022. In the dataset, 453,951 records are  ``laboratory-confirmed cases'', while 83,292 records are ``probable cases''. The records in the dataset are from 25 out of the 50 states in the US. The map of the data collection across states is shown in Figure \ref{fig:dataset}.

The remainder of this paper is organized as follows: the next section is the related works, followed by the proposed data-driven approach. 
Next comes the numerical results and discussions. Lastly, we have the conclusion. In this paper, ``nodes'', ``variables'', and ``features'' are used interchangeably; however, they all have the same meaning in context.


\section{Related Works}
The continued efforts in reducing the spread of the virus remains the responsibility of health agencies which develop vaccines (policies) and governments which fund (implement) them, respectively. 
Similarly, it is the responsibility of researchers to use theoretical expertise with experimental methodologies to facilitate new discoveries that can accelerate the early detection, prevention of the virus, and post-detection analysis.
From the research perspective, the availability of data and access to data are fundamental for both quantitative and qualitative analysis for knowledge extraction and/or model proposition. 
Most recent works which studied COVID-19 have employed a data-driven approach which leverages the power of artificial intelligence (AI) and ML. 
A subset of such works targets the identification and classification of COVID-19 cases based on known symptoms such as chills, fever, dry cough, and x-ray images \cite{abdul2021covid, liu2020experiments, wang2021covid}. 
The motivation behind this is to predict the chances of being infected with the disease when related symptoms are diagnosed in a patient.

To detect COVID-19 in people, the work in \cite{irawati2021classification} proposed the use of recorded cough sounds to train a XGBoost classifier algorithm. Cough sound features are extracted from the recording using the mel frequency cepstral coefficients (MFCC). 
The accuracy metric is used to evaluate the algorithm whose performance is 86.2\%. 
In a similar study in \cite{wang2021covid}, the health data of COVID-19 patients was used to train different ML models, such as Logistic Regression, Bagging Classifier, Balanced Bagging Classifier, Gradient Boosting Classifier, and XGBoost Classifier, with the best performing model selected to predict COVID-19 infection in patients. 
Considering the privacy of patients’ health data, a federated ML approach and a decentralized learning approach was adopted in \cite{abdul2021covid} and \cite{ajayi2023decentralized}, respectively, as against the traditional ML methods which requires accessing or transferring whole data to train the model. 
Specifically, federated ML uses distributed datasets rather than sharing raw data, and thus, provides solution for user data privacy, centralized computation, and transferred learning \cite{abdul2021covid, ajayi2023decentralized}. 
The federated ML paradigm in \cite{abdul2021covid} used a descriptive dataset and chest x-ray (CXR) images of COVID-19 patients to predict infectious cases and recovery rates. 
Another work in \cite{adetiba2022deepcovid} proposed an ML model for identifying COVID-19 virus sequences using genomic signal processing; the work compared the performance of the popular AlexNet model \cite{krizhevsky2017imagenet} and a heuristic convolutional neural network (CNN) model using Z-Curve images to train the models. The essence of COVID-19 infection detection (or identification) research is to facilitate the early diagnosis of patients’ disease and to avoid dangerous stages of the disease. 
Unlike the conventional clinical RT-PCR methods, the ML approach is noninvasive, and physicians can use patients’ symptoms or medical images for early diagnosis of the virus. 

Other related studies have paid attention to the stochastic behavior of the virus over time, that is, forecasting the spread of the virus using time-series data \cite{painuli2021forecast}. 
Several mathematical \cite{painuli2021forecast, ahmar2020suttearima, chintalapudi2020covid} and ML methods such as reinforcement learning (RL), recurrent neural networks (RNN) \cite{kumar2021recurrent} and deep learning have been proposed to predict the progression of COVID-19 cases. The autoregressive integrated moving average (ARIMA) method has been largely used for predictions on time-series data \cite{painuli2021forecast}, \cite{chintalapudi2020covid}. 
The ARIMA model was proposed in \cite{chintalapudi2020covid} to estimate registered and recovered cases after a lockdown in Italy, while the SutteARIMA model proposed in \cite{ahmar2020suttearima} was found to be more suitable than the ARIMA model for forecasting daily confirmed cases in Spain. 
The adaptive neuro-fuzzy inference system (ANFIS) based on the flower pollination algorithm (FPA) was used to forecast the number of confirmed cases of COVID-19 in China \cite{al2020optimization}. 
In the ML domain, long short-term memory (LSTM) network was used to forecast the future COVID-19 cases and to predict the possible stopping time of the outbreak in Canada \cite{chimmula2020time}. Polynomial neural network \cite{fong2020finding}, linear regression, multi-layer perceptron (MLP), and vector autoregression (VAR) have also been proposed in predicting the trend of COVID-19 cases in many countries. 
The essence of forecasting the trend of COVID-19 cases is to fore-tell the lifetime of the pandemic, and to decide timely and remedial actions.

While prior works have mostly targeted at the detection of COVID-19 infection in patients and forecasting of its spread using ML techniques, not many works have addressed and considered the post-detection analysis of COVID-19 to understand the virus and its threat to different demographics, in order to foster the extraction of accurate useful information to stop future resurgence of a similar disease outbreak. The work in \cite{pourbagheri2020laboratory} studied the importance of abnormal laboratory findings in COVID-19 diagnosis and prognosis, with the goal of offering essential assistance to discriminate between severe and non-severe COVID-19 cases. The method in the paper is mostly invasive, as it requires the collection of clinical features of patients' disease, such as complete blood count (CBC) to predict the COVID-19 prognosis. Although, the proposed laboratory method in \cite{pourbagheri2020laboratory} can be effective for clinicians and laboratory specialists, it does not serve the purpose of understanding the relationships among patients' symptoms from a non-invasive post-detection perspective which can serve the benefit of the health agencies, government, and the society. Considering the limitations in recent studies, this paper aims to give insight and reasoning into the causal relationships that exist among COVID-19 symptoms which is like the work in \cite{alsuwat2021detecting}. Although the method we propose is different from that of \cite{alsuwat2021detecting}, in that while \cite{alsuwat2021detecting} assumed a pre-known target variable to check the conditional dependencies, in this paper the probabilistic graph structure is learned from the COVID-19 dataset without any prior assertion on a target variable; instead the conditional dependencies for each variable is computed. On the other hand, we go further to discover the hidden knowledge in patients' symptoms as it relates to their demographics (i.e., age and gender) using an unsupervised ML algorithm. Lastly, to the best of the authors' knowledge, this paper is the first to use ML to facilitate the prediction of conditional probability distributions, where each class corresponds to a symptom class (as evidence) with demographics (as possible queries). This paper is an extension to the preliminary work in \cite{ajayi2023forecasting}.

\begin{table*}[htbp]
    \centering
        \caption{Directed acyclic graph (DAG) from the COVID-19 dataset. For example, in the Table we have $(B, H)=1$ and $(C, B)=1$, which means that there is an edge from node $B$ to node $H$ and there is an edge from node $C$ to node $B$ in the DAG respectively. The 24 features are represented as letters because of the limit of the Table width (see Table III for their original translation and meaning). }\label{tab:dag}
        \scriptsize
        \begin{tabular}{|c|c|c|c|c|c|c|c|c|c|c|c|c|c|c|c|c|c|c|c|c|c|c|c|c|} 
        \hline
         Features & A  & B  & C  & D  & E  & F  & G  & H  & I  & J  & K  & L  & M  & N  & O  & P  & Q  & R  & S  & T  & U  & V  & W  & X  \\ 
         \hline
         A & 0 & 0 & 0 & 0 & 0 & 0 & 0 & 0 & 0 & 0 & 0 & 0 & 0 & 0 & 0 & 0 & 0 & 0 & 0 & 0 & 0 & 0 & 0 & 0\\
         B & 0 & 0 & 0 & 0 & 0 & 0 & 0 & 1 & 0 & 0 & 0 & 0 & 1 & 1 & 0 & 1 & 0 & 1 & 1 & 0 & 0 & 1 & 0 & 0\\
         C & 0 & 1 & 0 & 0 & 0 & 0 & 0 & 0 & 0 & 0 & 0 & 0 & 1 & 1 & 1 & 0 & 0 & 0 & 1 & 0 & 0 & 0 & 0 & 0\\
         D & 0 & 0 & 0 & 0 & 1 & 1 & 1 & 0 & 1 & 0 & 1 & 0 & 0 & 0 & 0 & 1 & 1 & 0 & 0 & 1 & 0 & 0 & 0 & 1\\
         E & 1 & 0 & 0 & 0 & 0 & 1 & 0 & 0 & 1 & 1 & 0 & 0 & 0 & 0 & 0 & 0 & 1 & 0 & 0 & 0 & 1 & 0 & 1 & 1\\
         F & 0 & 0 & 0 & 0 & 0 & 0 & 0 & 0 & 1 & 1 & 0 & 0 & 0 & 0 & 0 & 0 & 0 & 0 & 0 & 0 & 1 & 0 & 0 & 1\\
         G & 0 & 0 & 0 & 0 & 0 & 0 & 0 & 0 & 0 & 0 & 0 & 0 & 0 & 0 & 0 & 0 & 0 & 0 & 0 & 0 & 0 & 0 & 0 & 0\\
         H & 1 & 0 & 0 & 1 & 1 & 1 & 0 & 0 & 0 & 0 & 0 & 1 & 0 & 0 & 0 & 1 & 1 & 0 & 0 & 0 & 0 & 0 & 1 & 0\\
         I & 0 & 0 & 0 & 0 & 0 & 0 & 0 & 0 & 0 & 1 & 0 & 0 & 0 & 0 & 0 & 0 & 0 & 0 & 0 & 0 & 1 & 0 & 0 & 0\\
         J & 0 & 0 & 0 & 0 & 0 & 0 & 0 & 0 & 0 & 0 & 0 & 0 & 0 & 0 & 0 & 0 & 0 & 0 & 0 & 0 & 0 & 0 & 0 & 0\\         
         K & 0 & 0 & 0 & 0 & 0 & 0 & 0 & 0 & 0 & 0 & 0 & 0 & 0 & 0 & 0 & 0 & 0 & 0 & 0 & 0 & 0 & 0 & 0 & 0\\         
         L & 1 & 0 & 0 & 1 & 1 & 1 & 1 & 0 & 1 & 1 & 0 & 0 & 0 & 0 & 0 & 0 & 1 & 0 & 0 & 0 & 1 & 0 & 1 & 1\\         
         M & 0 & 0 & 0 & 1 & 0 & 0 & 1 & 1 & 0 & 0 & 0 & 1 & 0 & 1 & 1 & 1 & 0 & 1 & 1 & 0 & 1 & 1 & 1 & 0\\         
         N & 0 & 0 & 0 & 0 & 0 & 0 & 0 & 0 & 0 & 0 & 0 & 0 & 0 & 0 & 1 & 0 & 0 & 0 & 1 & 0 & 0 & 1 & 0 & 0\\         
         O & 0 & 0 & 0 & 0 & 0 & 0 & 1 & 0 & 0 & 0 & 0 & 0 & 0 & 0 & 0 & 0 & 0 & 0 & 0 & 0 & 0 & 0 & 0 & 0\\         
         P & 0 & 0 & 0 & 0 & 0 & 0 & 0 & 0 & 0 & 0 & 0 & 0 & 0 & 0 & 0 & 0 & 0 & 0 & 0 & 1 & 1 & 0 & 1 & 0\\         
         Q & 1 & 0 & 0 & 0 & 0 & 1 & 0 & 0 & 1 & 1 & 0 & 0 & 0 & 0 & 0 & 0 & 0 & 0 & 0 & 0 & 1 & 0 & 1 & 1\\         
         R & 1 & 0 & 0 & 0 & 1 & 0 & 0 & 1 & 0 & 0 & 0 & 1 & 0 & 0 & 0 & 1 & 1 & 0 & 0 & 0 & 0 & 1 & 1 & 0\\         
         S & 0 & 0 & 0 & 1 & 1 & 0 & 0 & 1 & 1 & 0 & 0 & 0 & 0 & 0 & 1 & 0 & 0 & 1 & 0 & 0 & 0 & 0 & 1 & 0\\         
         T & 0 & 0 & 0 & 0 & 0 & 0 & 0 & 0 & 0 & 0 & 1 & 0 & 0 & 0 & 0 & 0 & 0 & 0 & 0 & 0 & 0 & 0 & 0 & 0\\         
         U & 0 & 0 & 0 & 0 & 0 & 0 & 0 & 0 & 0 & 0 & 0 & 0 & 0 & 0 & 0 & 0 & 0 & 0 & 0 & 1 & 0 & 0 & 0 & 0\\         
         V & 0 & 0 & 0 & 1 & 1 & 0 & 0 & 0 & 0 & 0 & 1 & 1 & 0 & 0 & 0 & 0 & 0 & 0 & 0 & 1 & 0 & 0 & 0 & 1\\         
         W & 1 & 0 & 0 & 0 & 0 & 1 & 0 & 0 & 0 & 1 & 0 & 0 & 0 & 0 & 0 & 0 & 0 & 0 & 0 & 0 & 0 & 0 & 0 & 0\\         
         X & 1 & 0 & 0 & 0 & 0 & 0 & 0 & 0 & 0 & 1 & 0 & 0 & 0 & 0 & 0 & 0 & 0 & 0 & 0 & 0 & 1 & 0 & 0 & 0\\         
        \hline        
        \end{tabular}
        \medskip
\end{table*}

\begin{table}[htbp]
    \centering
        \caption{Features in the COVID-19 dataset}\label{tab:dataset}
        \scriptsize
        \begin{tabular}{|l|l|l|l|l|} 
        \hline
         Code/Letter & Feature  & Type \\ 
         \hline
         A & abdominal pain & symptom \\
         B & abnormal chest X-ray & symptom \\
         C & \multicolumn{1}{p{70pt}|}{acute respiratory distress syndrome} & symptom \\
         D & age group & demographic \\
         E & chills & symptom \\
         F & cough & symptom \\
         G & death & severity \\
         H & diarrhea & symptom \\
         I & fever & symptom \\
         J & subjective fever & symptom \\
         K & health care worker & profile \\
         L & headache & symptom \\
         M & hospitalized & severity \\
         N & \multicolumn{1}{p{70pt}|}{intensive care unit (ICU)} & severity \\
         O & mechanical ventilation & severity \\
         P & medical condition & severity \\
         Q & muscle aches (myalgia) & symptom \\
         R & nausea or vomiting & symptom \\
         S & pneumonia & symptom \\
         T & race & demographic \\
         U & runny nose & symptom \\
         V & gender & demographic \\
         W & shortness of breath & symptom \\
         X & sore throat & symptom \\         
         \hline         
        \end{tabular}
        \medskip
\end{table}

\section{Proposed Data-driven Approach}
In this section, the proposed three-stage data-driven method for explaining the severity of COVID-19 are presented.
The section begins with the definition of probabilistic graphical models (PGMs) with Bayesian networks, followed by the clustering of data with ML, and the design of the demographic symptom identification (DSID) model. Lastly in this section, we describe the COVID-19 dataset and its features.

\subsection{Probabilistic Graphical Models}
Most decision-making tasks require a person to reason, and such reasoning is mostly based on either prior experience or current evidence. For example, a medical doctor needs to take patients’ symptoms, diagnosis, and perhaps personal characteristics to determine what disease the patient has and the possible treatments to administer \cite{koller2009probabilistic}. The complexity of decision-making in a challenging situation such as disease diagnosis requires a general framework or model that can encode knowledge from observations and answer several questions of uncertainty. This is because, the world itself is rarely deterministic by limited observations, and thus to obtain meaningful conclusions, we need to reason about what is probable rather than what is possible \cite{koller2009probabilistic}. 

The knowledge of probability theory provides the framework for assessing multiple variables and their likelihood; this makes the modeling of complex reasoning more faithful to reality. 
The reasoning task is to estimate the probability of one or more variables given the observation of other variables. 
The joint probability distribution over the space of possible states (where each variable has one or multiple states) for the set of random variables $\mathcal{V}$ need to be constructed. 
However, computing the joint distribution over a set of dozens or hundreds of random variables is a daunting task, but PGMs can be used to simplify the construction of complex distributions. 
PGMs as the name implies use a graph-based representation to encode the conditional independencies between random variables. The graph depicts the causal relationships that exist between variables in the distribution. 

Consider a distribution modeled as a graph $\mathcal{G}(\mathcal{V}, \mathcal{E})$ where $\mathcal{V}$ is the set of variables and $\mathcal{E}$ is the set of edges; the variable interactions can be expressed in the form $(A \perp B \mid C)$, where $A, B, C \in \mathcal{V}$ and $(C,A) \in \mathcal{E}$.
The joint probability distribution over a set of variables $\mathcal{V} = \{V_1, …, V_n\}$ can be expressed as $P(V_1 = v^k_1, …, V_n = v^k_n)$ where $v^k_i \in \mathcal{S}_i$ and $\mathcal{S}_i$ is the set of variable states for variable $V_i$. 
With the graph structure, the joint distribution can be decomposed into smaller factors based on the conditional independencies among variables. 
The joint distribution is defined as a product of the factors as:

\begin{equation}
P(V_1, …, V_n) = \prod_{i=1}^{n}P(V_i \mid Pa(\hat{V_i}))
\end{equation}

\begin{equation}
\sum_{v^k_i \in \mathcal{S}_i} P(V_i=v^k_i \mid Pa(\hat{V_i})) = 1
\end{equation}
where $n$ is the total number of variables in $\mathcal{G}$, while $V_i$ and $Pa(\hat{V_i})$ represent a variable in $\mathcal{V}$ and the ``parents'' of the variable respectively. This is also referred to as the conditional probability distribution of the variable set. By rule, the probability distribution of a variable over its possible states, given (or conditioned on) its evidence variables (as parents with their defined states) should sum to 1.

\subsubsection{Bayesian Networks}
In implementing the graphical representation of distributions there are two main approaches. 
One is the Bayesian networks in which case the edges are directed as $parent \rightarrow child$, and the other is Markov networks in which case the edges are undirected. 
Both networks differ in the set of independencies they encode and in the induced factorization from the distribution.
For example, consider a three-variable Bayesian network $A \leftarrow C \rightarrow B$ with the independence assertion that  $(A \perp B \mid C)$, where $A$ is a ``child'' of $C$ and is independent of $B$.
This can be interpreted as the Markov condition \cite{tsagris2019bayesian} which states that \textit{each variable is independent of its non-descendants given its parent(s)}; in this example, $B$ is not a descendant of $A$, and the edge $C \rightarrow A$ exists if $C$ is a direct cause of $A$.
Probabilistically, the conditional independence holds if $P(A,B \mid C)=P(A \mid C)P(B \mid C)$.

Formally, a Bayesian network or BN for short \cite{pearl1988probabilistic}, is a directed acyclic graph (DAG) $\mathcal{G}$ over variables $V_i \in \mathcal{V}$ and a joint probability distribution $P$. The directed graph is acyclic because no cycles are allowed. From a BN, any probability of the form $P(V_i \mid Pa(\hat{V_i}))$ can be determined, i.e., the conditional probability on each variable can be computed. 
Note that if a variable has no parent, only the marginal probability $P(V_i)$ is computed. 
In practice, a BN structure is constructed from a dataset, and once a feasible structure is found, we can run inference on the dataset. In this paper, we leverage the BN to estimate a DAG that captures the dependencies between the variables in the COVID-19 dataset.

\subsubsection{Structure and Parameter Learning from Dataset}
Constructing a BN requires learning the conditional independencies in the dataset; there are several structure learning algorithms which search the best graphical structure for the dataset. 
There are two main structure learning techniques – score-based structure learning and constraint-based structure learning. 
For both techniques, a learning algorithm traverses the search space of possible structures and selects the one with the optimal score. The most popular ones of these algorithms include the Exhaustive search, HillClimb search, PC algorithm \cite{tsagris2019bayesian} and MMHC algorithm \cite{tsamardinos2006max}.

The search space of DAGs is always super-exponential with respect to the number of variables in the dataset. 
The exhaustive search algorithm always attempts to identify the ideal structure, but only if the dataset consists of no more than five features. 
Since this is not realistic in practice, as we need to handle high-dimensional data, a heuristic search algorithm helps to find a good structure that implements a greedy local search and terminates when a local maximum is reached. 
We implemented the Hill-Climb search algorithm as a greedy local search to construct the DAG from the COVID-19 dataset. Each sample in the dataset has 24 features; the algorithm thus returns a DAG with 24 nodes and 109 edges. The output of the algorithm is usually a figure (or plot) representing the set of tuples (or edges) that captures the (in)dependencies among the variables, however it is strenuous and difficult for anyone to easily interpret the DAG from the figure, owing to the large number of edges. We therefore resort to Table \ref{tab:dag} which shows the DAG from the dataset. Each element $(i, j)$ in Table \ref{tab:dag} which corresponds to the edge (or link) from node $i$ to node $j$ is defined as follows:

$$
    (i, j) = 
    \begin{cases}
    1     &\text{if   } (i, j) \in \mathcal{E} \text{ and } i \ne j \\
    0     &\text{otherwise} \\
    \end{cases}
$$

When the most probable network structure has been found, the important task is to estimate the conditional probability distributions (CPDs), which is defined as the set of parameter values $\theta_\mathcal{G}$ for the variables $V_i \in \mathcal{V}$.
For each variable $V_i$, there is a separate multinomial probability distribution over every possible parent’s instantiation with values $v^k_i$; thus each variable’s CPD is considered a full table. 
Estimating the CPDs for each variable naturally follows the use of relative frequencies with respect to the occurrence of the variable states; this approach is called maximum likelihood estimation (MLE). 
The MLE has the problem of overfitting the data when there are not enough observations in the data. 
In contrast, the Bayesian parameter estimator starts with some pseudo state counts before observing the data; this reduces the tendency of overfitting or biased estimation.

The method used for learning the parameters, i.e., estimating the CPDs for each variable in the COVID-19 dataset, is the Bayesian parameter estimation method. 
The MLE which uses the relative frequencies is considered not to benefit the objective of optimal CPDs.

\subsection{Data Clustering}
Clustering of data is important in learning the internal attributes of the data or inherent patterns that cannot be humanly detected, because it is often impossible and laborious to superficially identify similarities between high-dimensional data. 
The goal of clustering is to find subgroups in the dataset in a manner that data points which have significant similar properties belong to the same cluster; this as well translates to that, data points that are not similar belong to different clusters. 
In clustering algorithms, the Euclidean-distance function is used as a metric to determine the assignment of data points to appropriate clusters or class. 
Unlike supervised learning where data points are mapped to their target labels, clustering is an unsupervised learning task since no one knows the ground truth label to compare and evaluate with the outcomes from clustering \cite{kmeans}.

The most popular and widely used clustering algorithm is the Kmeans algorithm \cite{lloyd1982least}, \cite{liu2020determine} due to its simplicity and robustness in fitting data points. 
As the name suggests, the Kmeans algorithm partitions the dataset into $K$ pre-defined classes where each data point belongs to only one class. 
The clustering algorithm tries to minimize an objective function which is the sum of squared error (SSE) between the data points and the cluster centers as follows:

\begin{equation}
\underset{\mu_1,\cdots,\mu_K \in Z}{\text{minimize}} \quad \sum_{k=1}^K \sum_{z \in Z_k} \lVert z - \mu_k \rVert^2,
\end{equation}
where $Z_k$ is the set of data points that are closest to center point $\mu_k$ more than other data points.

Specifically, the algorithm tries to minimize the intra-cluster distance and maximize the inter-cluster distance, while keeping the SSE as low as possible. 
The Kmeans algorithm works by initializing $K$ centroids (i.e., cluster centers) randomly and assigning data points to the centroids, this continues to iterate until there is no change in the assignment of data points to centroids, in which case, the algorithm converges. 
A drawback with this algorithm is that, since the initialization of centroids is random, the algorithm may be stuck in a local minimum and not reach the global minimum, as different initializations usually lead to different clusters. 
This problem has been solved with the Kmeans++ algorithm \cite{arthur2007k} which works similarly as the Kmeans algorithm but differs in the centroid initialization process; centroids are initialized smartly, and the quality of clustering is improved.

In our setting, the DAG obtained from the COVID-19 dataset provides a guide in selecting symptoms that have correlation with the age and gender of patients; such relationship cannot be accurately determined by a human expert or health practitioner with domain knowledge, as patients are usually sampled disparately. Let $\mathcal{F}$ denote the set of features in the dataset, $\mathcal{F}_t$ denote the set of demographic (target) variables and $\mathcal{F}_c \subset \mathcal{F}$ denote the set of features to be used for clustering, where $\mathcal{F}_c= \{ f \mid f \in \textit{Ancestor}(\mathcal{F}_t), \forall f \in \mathcal{F} \}$.
For clarity, a feature $f$ is an \textit{Ancestor} of another feature $f_t \in \mathcal{F}_t$ if $f$ belongs to the incoming chain into $f_t$. On the other hand, a feature $f$ is a \textit{Descendant} of another feature $f_t \in \mathcal{F}_t$ if $f$ is a member of the outgoing chain from $f_t$. 

Note that the output from clustering are the cluster labels for each $Z_k \in Z$, which will then be used as labels to train the DSID model in a supervised learning manner.

\subsection{Demographic Symptom Identification Model}
The task of predicting the demographic symptom class of patients from disease symptoms is non-trivial, and thus requires an intelligent and computationally efficient method. Considering the multiple target variables for this task, the feature selection strategy will traditionally require significant domain knowledge, and even with that, there is no guarantee of optimum prediction. This is because there can be strong correlation between some features and a target variable, while those features may be weakly correlated with another target variable. One method to address the multi-variate classification task, is to train a separate ML classifier with those features that are highly correlated with each of the target variables. This means that for $T$ target variables in the dataset, $T$ number of classifiers will be trained, and the computational complexity will be $O(NMT)$ where 
$M$ is the number of training samples and $N$ is the dimension of features in the dataset. Another method to solve the problem is to design a multi-output classifier $g(\textbf{x}):\mathbb{R}^N \rightarrow \{f_t\}^T_{t=1}$ that maps the input \textbf{x} to each of the target demographic variables $f_t$, that is, multi-label mapping, where $f_t=\{c_1, \cdots, c_k\}$ and $c_k$ is the $k$-th class or possible state of $f_t$. This also has a computational complexity of $O(NMT)$ and will require more layers and parameters to train the model.

However, we can leverage the DAG obtained from the BN structure learning in stage 1 shown in Figure \ref{fig:main}, to select features that are highly correlated with the demographic variables $\mathcal{F}_t$. This will reduce the computational complexity to only $O(NM)$. We design the DSID model to train on the dataset in a supervised learning manner with labels obtained from clustering as shown in stage 2 and stage 3 of Figure \ref{fig:main}.

\subsection{COVID-19 Dataset Features }
The COVID-19 dataset consists a total 24 features which can be sub-divided into: 16 disease features, 3 demographic features, 4 severity features and 1 patient-profile feature. We clarify that the disease features correspond to the COVID-19 symptoms, while the severity features refer to the criticalness of a patient's condition; these include hospitalization, admittance into intensive care unit (ICU), the use of mechanical ventilation, and death of the patient. Table \ref{tab:dataset} shows each feature description with the use of letters of the English alphabet for ease of representation.
All of the features in the dataset except race, age, and gender, have two possible states, ``yes'' and `no''.

We encourage the reader to consider Table \ref{tab:dataset} as a lookup or reference table to be used when interpreting the CPDs of each of the severity variables in Section \ref{sec:results}.

\begin{figure}
    \centering
    \includegraphics[width=0.5\textwidth]{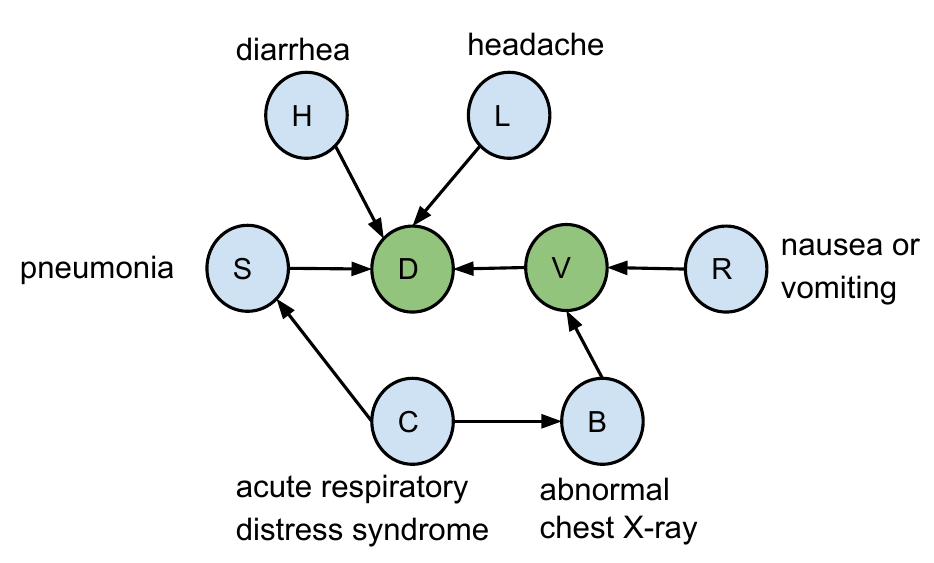}
    \caption{DAG showing relationship between predictor variables $\mathcal{F}_c$ and target variables $\mathcal{F}_t$ to facilitate features selection to train the DSID model.}\label{fig:subDAG}
\end{figure}

\begin{table}[ht]
\caption{DSID model parameters and hyperparameters}\label{modelparams}
    \centering
    \scriptsize
        \begin{tabular}{|l|c|} 
             \hline
             Name & Value \\ 
             \hline
             Epochs &  16\\ 
             Batch size &  50\\              
             Layers & 6\\ 
             Optimizer & Adam\\
             Regularization & Early stopping\\
             $\mathbf{W_1},\mathbf{W_2}$ & $128\times6$, $128\times128$\\
             $\mathbf{W_3}, \mathbf{W_4}$ & $64\times128$, $64\times64$\\
             $\mathbf{W_5}, \mathbf{W_6}$ & $32\times64$, $27\times32$\\
             \hline
            \end{tabular}
            \medskip
    \label{tab:param}
\end{table}

\section{Numerical Results and Discussions}
\label{sec:results}
In this section, we present the outcome of the parameter learning of the BN from the dataset, and the performance of the DSID model. Note that while the DAG captures the dependencies between variables in the dataset, the parameter learning reveals and estimates the conditional or marginal probability distributions of the individual variables.

\subsection{Experiment Settings}
For the clustering stage, we computed the DAG that captures the relationship between $\mathcal{F}_c$ and $\mathcal{F}_t$, which are colored blue and green respectively, as shown in Figure \ref{fig:subDAG}. It can also be inferred that this is a sub-DAG from the DAG presented in Table \ref{tab:dag}. Since we do not have knowledge about the patients' symptoms similarities in the dataset, it is inefficient to randomly guess the optimal $K$ value a-priori for clustering the data. Hence, we used the popular Dunn's index \cite{ilc2012modified}, \cite{sinaga2020unsupervised} algorithm, which computes the optimal number of clusters that maximizes the ratio between the minimum distance between two clusters and the size of the largest cluster. The returned $K$ value from the Dunn's index algorithm in our experiment is $K=27$, which is then used for clustering the data with the Kmeans++ algorithm.

For the patients' symptoms demographic prediction, the DSID model architecture with fully-connected (FC) layers is summarized as:
\begin{align}\label{eq:AID}
    &\text{FC}_i(\mathbf{x}) \triangleq \text{ReLU}(\mathbf{W}_i \mathbf{x}+\mathbf{b}_i)\\
    &\hat{\mathbf{y}} = 
    \text{softmax}(\mathbf{W}_6(\text{FC}_5\cdots(\text{FC}_1(\mathbf{x}))+\mathbf{b}_6)
\end{align}
where \textbf{W} and \textbf{b} are the weight matrix and bias vector, respectively, in the hidden layers and output layer. Table \ref{tab:param} lists the configurations of hyperparameters used for training the DSID model. 

The \textit{categorical cross-entropy} loss function is used for back propagation to optimize the model's performance through weight updates in each layer. It is defined as:

\begin{equation}
    \mathcal{L}(\mathbf{X_\mathcal{B}}, \mathbf{y}, \mathbf{W}) = - \frac{1}{|\mathcal{B}|} \sum_{\mathbf{x_i} \in \mathbf{X_\mathcal{B}}} \sum_{k=1}^K y_{ik}\text{ log }\hat{y}_{ik}
\end{equation}
where $\mathbf{X_\mathcal{B}}$, $\mathbf{y}$ and $|\text{ . }|$ denote the instances in batch $\mathcal{B}$, the labels of the instances, and cardinality of the set $\mathcal{B}$, respectively.

Our experiments are run on the Google Colab cloud environment, equipped with Nvidia K80 GPUs and Intel Xeon processors @ 2.3GHz. We implemented the Bayesian networks using the pgmpy python package. Other packages used are Numpy \cite{harris2020array} and Tensorflow \cite{girija2016tensorflow}. The codes for the experiment are publicly made available on \href{https://github.com/Seunaj/Covid-19-Bayesian-Networks-CPDs}{https://github.com/Seunaj/Covid-19-Bayesian-Networks-CPDs}.

\subsection{Severity Explanation}
\begin{figure}
    \centering
    \includegraphics[width=0.5\textwidth]{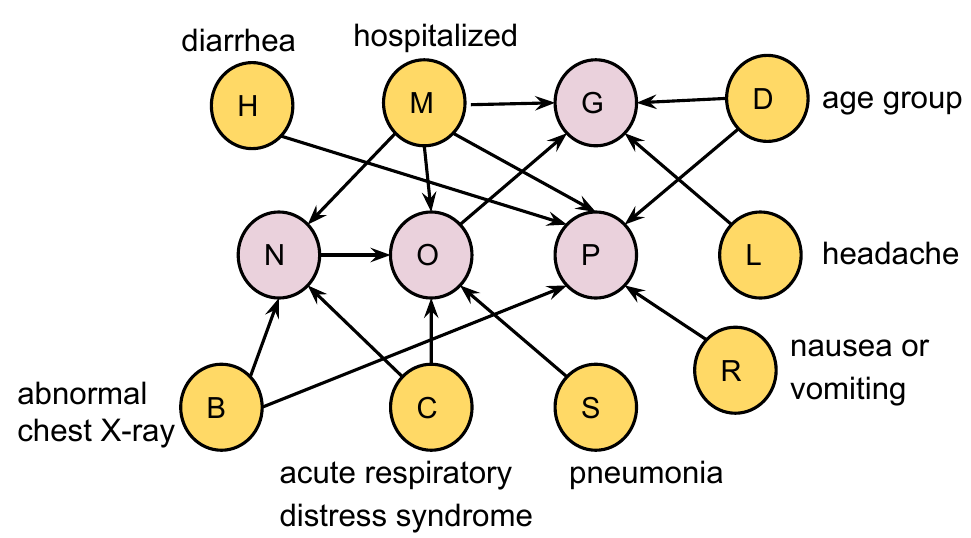}
    \caption{DAG showing conditional dependencies between severity variables (G, N, O, P) in color purple and their parents in color orange.}\label{fig:severity_subDAG}
\end{figure}

The CPDs of each variable is useful to understand how much the variables depend and affect each other. With the local independencies in the DAG, we can significantly reduce the number of estimated parameters. However, while most variables have two possible states, some others such as race and age have six and nine possible states, respectively. Thus, if any of these two variables is a parent of (or affects) another variable, the parameter space will be large owing to the different possible parent-to-child configurations, but estimating the CPDs is still computationally efficient compared to computing the joint probability distribution on all the variables. To facilitate a reader's understanding of the CPDs, we use Figure \ref{fig:severity_subDAG} to illustrate the conditional dependencies between the severity variables and their evidences (or parents).

Because of space limitation, we will only focus on the CPDs of the severity variables: death (G), ICU (N), mechanical ventilation (O), and medical condition (P). These CPDs will be used to explain the COVID-19 cases in the dataset, and serve to give insightful direction on possible decisions that can facilitate the awareness of the public and health agencies in addressing the spread and threats.

\subsubsection{CPD for Intensive Care Unit}
    The ICU (denoted as N), as widely known, is a unit in a hospital where a patient who is in critical or life-threatening condition is catered for, under close monitoring with life-support equipment. The severity of diseases such as COVID-19 can necessitate a patient's admittance to the ICU. From the DAG of the BN in Table \ref{tab:dag}, we can see that the decision on whether to take a patient to the ICU or not, depends on: abnormal chest X-ray (B), acute respiratory distress syndrome (C) and hospitalized (M). Thus, the CPD for ICU can be defined as $P(N \mid B, C, M)$. From Table \ref{tab:icu} which shows the CPD for the ICU variable, we can learn the following observations:
\begin{itemize}
    \item The probability of not taking a patient to the ICU is 1 if there is no evidence of abnormal chest X-ray and acute respiratory distress syndrome, and if the patient is not hospitalized. This clearly means that the patient is not in a critical condition.
    \item The probability that a patient should be taken to the ICU is 0.54 if there is evidence that the patient has both abnormal chest X-ray and acute respiratory distress syndrome, and that the patient is already hospitalized.
\end{itemize}

The CPD provides a clear and meaningful approach that can assist medical practitioners, e.g., doctors and nurses, to make a decision on the admittance of a patient to the ICU by only examining variables B, C, and M.

\subsubsection{CPD for Mechanical Ventilation}
The mechanical ventilation (denoted as O), as widely known, is the use of a mechanical ventilator to deliver oxygen into the lungs and remove carbon-dioxide out of the lungs. When a patient's condition is critical or severe (with lungs collapsing), the patient is taken to the ICU and assisted with the ventilator. From the DAG of the BN in Table \ref{tab:dag}, the decision to use a mechanical ventilator for a patient depends on: acute respiratory distress syndrome (C), hospitalized (M), ICU (N) and pneumonia (S). Thus, the CPD for mechanical ventilation can be defined as $P(O \mid C, M, N, S)$. From the result in Table \ref{tab:mechvent}, we can make some statements, as follows:
\begin{itemize}
    \item The likelihood of using a mechanical ventilator to facilitate a patient's respiration is 0.43, if the patient has only acute respiratory distress syndrome, and is hospitalized and in the ICU. However, should the patient's condition be complicated with pneumonia, the probability of using a mechanical ventilation is increased to 0.63.
    \item There is no need to use a mechanical ventilation if the patient is never hospitalized and not in the ICU; the probability is $\approx$ 1 in this case.
\end{itemize}

One important thing to note is that the probability of using the mechanical ventilator increases with respect to the degree of the patient's ill-medical condition, featuring symptoms like pneumonia and acute respiratory distress syndrome.

\subsubsection{CPD for Death}
\begin{figure*}
\centering
  \includegraphics[width=\textwidth]{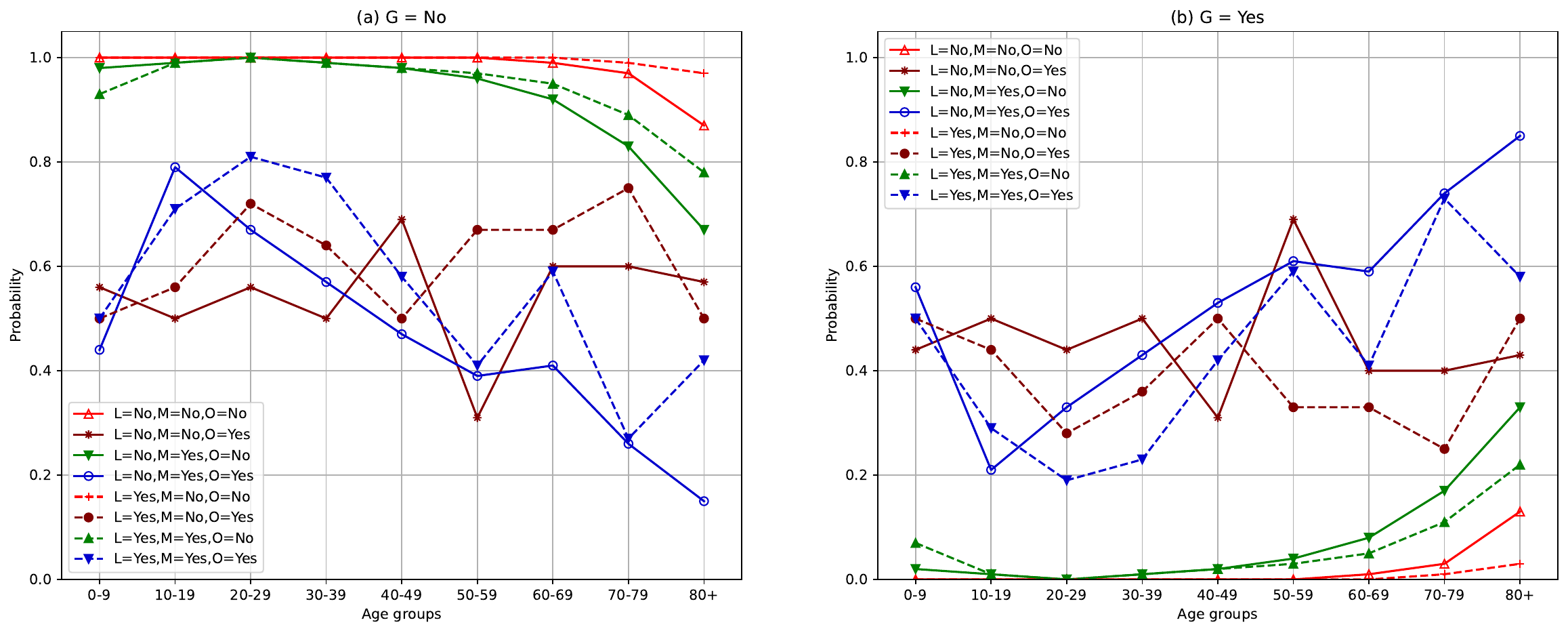}
  \caption{Conditional probability distribution showing how the causal variables (see legend map) cause death across different age-groups (x-axis). The subplot in (a) represents the probability for when death = No, and the subplot in (b) is probability for when death = Yes.}\label{fig:CPD_Death}
\end{figure*}

The death variable (denoted as G), is by far the most examinable severity variable as it is important to help us know how to reduce the death rate due to COVID-19, and to prevent possible future resurgence of the disease. Captured by the DAG in Table \ref{tab:dag}, whether a patient will die by COVID-19 or not, depends on: age group (D), headache (L), hospitalized (M) and mechanical ventilation (O). It is interesting to observe from our DAG that the age group that a patient belongs to affects the death probability of the patient, and least expected (e.g., to a non-medical expert) is that headache is an important symptom that should be checked in a patient that is infected with COVID-19. The CPD for death can be defined as $P(G \mid D, L, M, O)$, and from Figure \ref{fig:CPD_Death} we can infer that:

\begin{itemize}
    \item Across all the age groups, there is always a high probability of death if a patient uses the mechanical ventilator as a breathing aid. However, the probability is higher for ages between 0-9 years, and ages above 50 years. For example, the probability that a patient may die is greater than 0.7 if the patient is under mechanical ventilation and is above 70 years.
    \item There is a high probability (up to 0.81) of survival for ages below 50 years who suffer from COVID-19 and use the mechanical ventilation.
    \item There is stochasticity in the observation where a patient is not hospitalized, but uses a mechanical ventilation. We do not want the reader to be confused, should the outcome not match basic human intuition.
    \item The impact of headache on causing COVID-19 death is negligible across ages below 70 years, but it can cause death with high probability for ages above 70 years, even without the use of mechanical ventilation as long as the patient is hospitalized.
    \item As expected, across the majority of the age groups, the probability that a patient will survive is $\approx$ 1, if there is no evidence of headache, being hospitalized, or use of mechanical ventilation. Although for patients above 80 years the probability is 0.87. This can be as a result of other underlying medical conditions.
    \item In general, children and youths (i.e., ages 40 and below) have a higher survival rate than older people, with respect to the the severity of the disease.
\end{itemize}

From our results and discussions so far, we can assert that critical COVID-19 symptoms such as pneumonia (S) and acute respiratory distress syndrome (C) are the causes of hospitalized (M) and mechanical ventilation (O), which then cause death.

\subsubsection{CPD for Medical Condition}
\begin{figure*}
\centering
  \includegraphics[width=\textwidth]{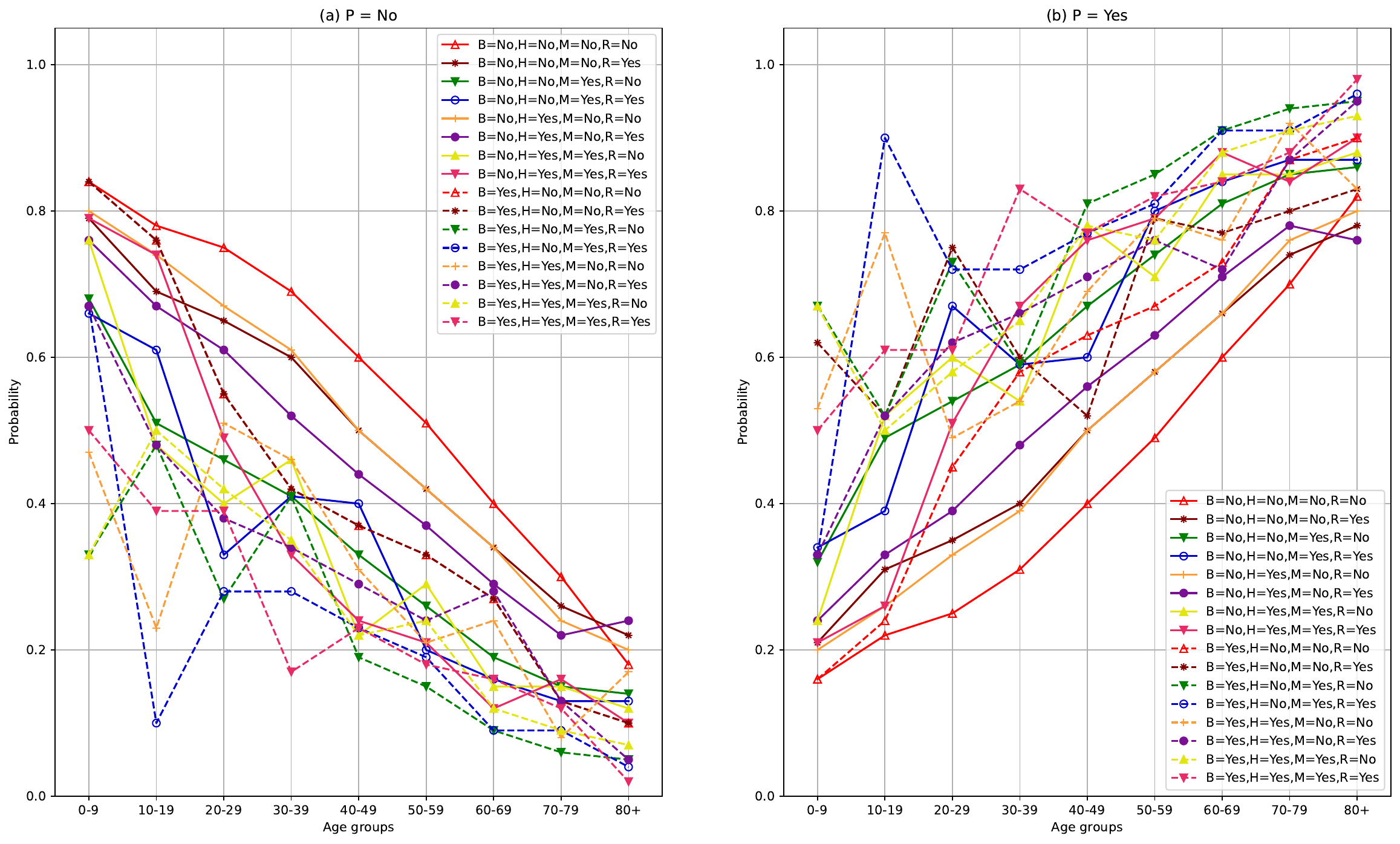}
  \caption{Conditional probability distribution of patients' medical condition showing the variation across different age groups (x-axis) and causal variables (see legend map). The subplot in (a) represents the probability for when a patient has no underlying disease, i.e., medical condition = No, and the subplot in (b) is probability for when medical condition = Yes.}\label{fig:CPD_MedCondition}
\end{figure*}

The severity of COVID-19 in a patient can be largely due to some underlying medical conditions (denoted as P). However, we cannot easily predict (or guess) what variables to examine in order to validate this claim or assumption. The DAG in Table \ref{tab:dag} shows that medical condition depends on: abnormal chest X-ray (B), age group (D), diarrhea (H), hospitalized (M) and nausea vomiting (R). This means that we can estimate the likelihood of a patient having a medical condition, given the evidence variables, that is, finding $P(P \mid B, D, H, M, R)$. Recall that the cases in the dataset are all positive cases of COVID-19, so we are not examining, as a-priori, if having underlying medical conditions will cause COVID-19 or not. Instead we want to know, as a-posterior, whether patients who are diagnosed of COVID-19 had underlying medical conditions or not. From Figure \ref{fig:CPD_MedCondition} we can infer that:

\begin{itemize}
    \item For all the age groups, patients that are hospitalized have a high probability of suffering from underlying medical conditions, and this is more evident in the categories of 20 years and beyond, with the worst case being for patients who are 50 years and beyond. The risk of having medical conditions is even stronger if a patient is diagnosed of abnormal chest X-ray (up to 0.98 for 80 years and above).
    \item The effect of diarrhea and nausea vomiting as causes of underlying medical condition is low compared to abnormal chest X-ray, except for patients above 50 years, where these symptoms contribute to their probability outcomes, with nausea vomiting being the larger contributor.
    \item In general, there is a low probability for children within the ages of 0 - 19 years to suffer from underlying medical conditions, except if diagnosed of abnormal chest X-ray. This is unlike the case for the older population.
\end{itemize}

The analysis of CPDs of the severity variables G, N, O, and P, are not limited to what we have discussed in this paper. There are more richer information in the results that the reader should extract and learn from. We have only provided a guide on how to interpret the conditional dependencies in the dataset, the probability distributions and useful information that a non-expert reader can benefit.


\begin{table*}[!htbp]
\caption{{ICU (N)} }
\label{tab:icu}
\def\arraystretch{1.2}
\ignorespaces 
\centering 
\scriptsize
\begin{tabulary}{\linewidth}{|p{\dimexpr.1029\linewidth-2\tabcolsep}|p{\dimexpr.1171\linewidth-2\tabcolsep}p{\dimexpr.11\linewidth-2\tabcolsep}p{\dimexpr.11\linewidth-2\tabcolsep}p{\dimexpr.11\linewidth-2\tabcolsep}p{\dimexpr.11\linewidth-2\tabcolsep}p{\dimexpr.11\linewidth-2\tabcolsep}p{\dimexpr.11\linewidth-2\tabcolsep}p{\dimexpr.12\linewidth-2\tabcolsep}|}
\hline
& B (No) & B (No) & B (No) & B (No) & B (Yes) & B (Yes) & B (Yes) & B (Yes)\\ 
\hline
& C (No) & C (No) & C (Yes) & C (Yes) & C (No) & C (No) & C (Yes) & C (Yes)\\
\hline
& M (No) & M (Yes) & M (No) & M (Yes) & M (No) & M (Yes) & M (No) & M (Yes)\\
\hline
N (No) & 1.0 & 0.91 & 0.95 & 0.61 & 0.99 & 0.8 & 0.82 & 0.46\\
N (Yes) & 0.0 & 0.09 & 0.05 & 0.39 & 0.01 & 0.2 & 0.18 & 0.54\\ 
\hline
\end{tabulary}
\medskip
\end{table*}

\begin{table*}[!htbp]
\caption{{Mechanical ventilation (O)} }
\label{tab:mechvent}
\def\arraystretch{1.2}
\ignorespaces 
\centering 
\scriptsize
\begin{tabulary}{\linewidth}{|p{\dimexpr.1029\linewidth-2\tabcolsep}|p{\dimexpr.1171\linewidth-2\tabcolsep}p{\dimexpr.11\linewidth-2\tabcolsep}p{\dimexpr.11\linewidth-2\tabcolsep}p{\dimexpr.11\linewidth-2\tabcolsep}p{\dimexpr.11\linewidth-2\tabcolsep}p{\dimexpr.11\linewidth-2\tabcolsep}p{\dimexpr.11\linewidth-2\tabcolsep}p{\dimexpr.12\linewidth-2\tabcolsep}|}
\hline
& C (No) & C (No) & C (No) & C (No) & C (No) & C (No) & C (No) & C (No)\\ 
\hline
& M (No) & M (No) & M (No) & M (No) & M (Yes) & M (Yes) & M (Yes) & M (Yes)\\
\hline
& N (No) & N (No) & N (Yes) & N (Yes) & N (No) & N (No) & N (Yes) & N (Yes)\\
\hline
& S (No) & S (Yes) & S (No) & S (Yes) & S (No) & S (Yes) & S (No) & S (Yes)\\
\hline
O (No) & 1.0 & 1.0 & 0.86 & 0.67 & 0.99 & 0.99 & 0.85 & 0.76\\
O (Yes) & 0.0 & 0.0 & 0.14 & 0.33 & 0.01 & 0.01 & 0.15 & 0.24\\
\hline
\end{tabulary}
\medskip

\scriptsize
\begin{tabulary}{\linewidth}{|p{\dimexpr.1029\linewidth-2\tabcolsep}|p{\dimexpr.1171\linewidth-2\tabcolsep}p{\dimexpr.11\linewidth-2\tabcolsep}p{\dimexpr.11\linewidth-2\tabcolsep}p{\dimexpr.11\linewidth-2\tabcolsep}p{\dimexpr.11\linewidth-2\tabcolsep}p{\dimexpr.11\linewidth-2\tabcolsep}p{\dimexpr.11\linewidth-2\tabcolsep}p{\dimexpr.12\linewidth-2\tabcolsep}|}
\hline
& C (Yes) & C (Yes) & C (Yes) & C (Yes) & C (Yes) & C (Yes) & C (Yes) & C (Yes)\\ 
\hline
& M (No) & M (No) & M (No) & M (No) & M (Yes) & M (Yes) & M (Yes) & M (Yes)\\
\hline
& N (No) & N (No) & N (Yes) & N (Yes) & N (No) & N (No) & N (Yes) & N (Yes)\\
\hline
& S (No) & S (Yes) & S (No) & S (Yes) & S (No) & S (Yes) & S (No) & S (Yes)\\
\hline
O (No) & 0.97 & 0.93 & 0.5 & 0.71 & 0.94 & 0.96 & 0.57 & 0.37\\
O (Yes) & 0.03 & 0.07 & 0.5 & 0.29 & 0.06 & 0.04 & 0.43 & 0.63\\
\hline
\end{tabulary}
\medskip
\end{table*}

\begin{figure}
    \centering
    \includegraphics[width=0.5\textwidth]{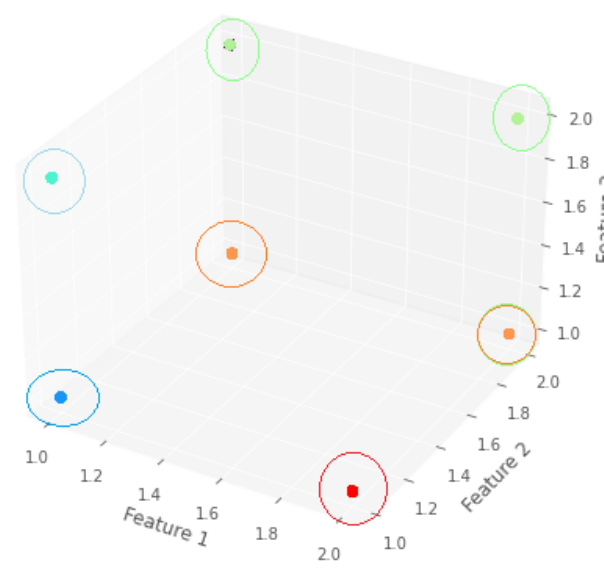}
    \caption{Illustration of the quality of Kmeans++ clustering. For ease of visualization, we give the plot in 3-D with each axis corresponding to the top three features. We only show clustering for seven classes here.}\label{fig:cluster}
\end{figure}

\begin{figure}
    \centering
    \includegraphics[width=0.5\textwidth]{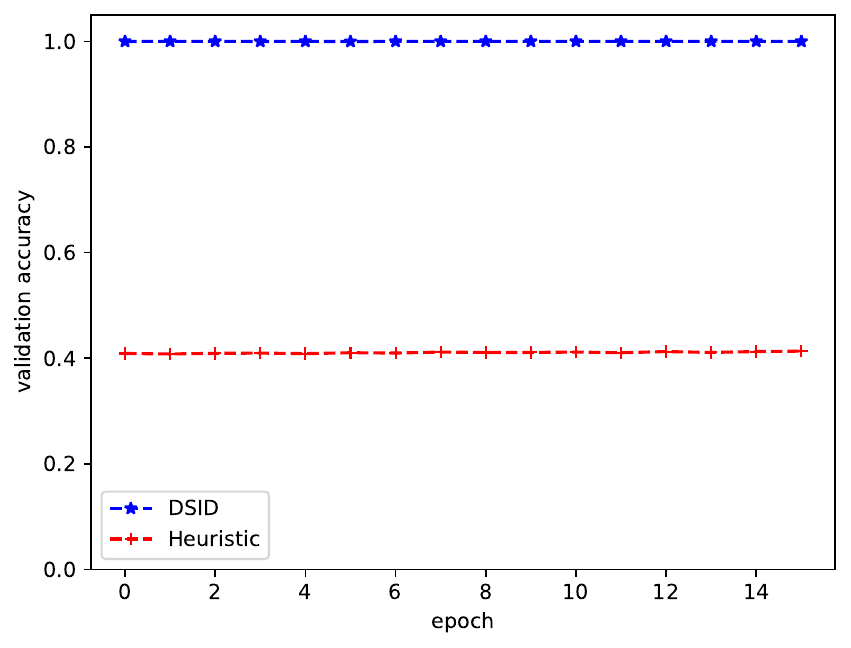}
    \caption{Performance comparison on validation set accuracy between DSID model and a heuristic ML method.}\label{fig:validation}
\end{figure}

\begin{figure}
    \centering
    \includegraphics[width=0.5\textwidth]{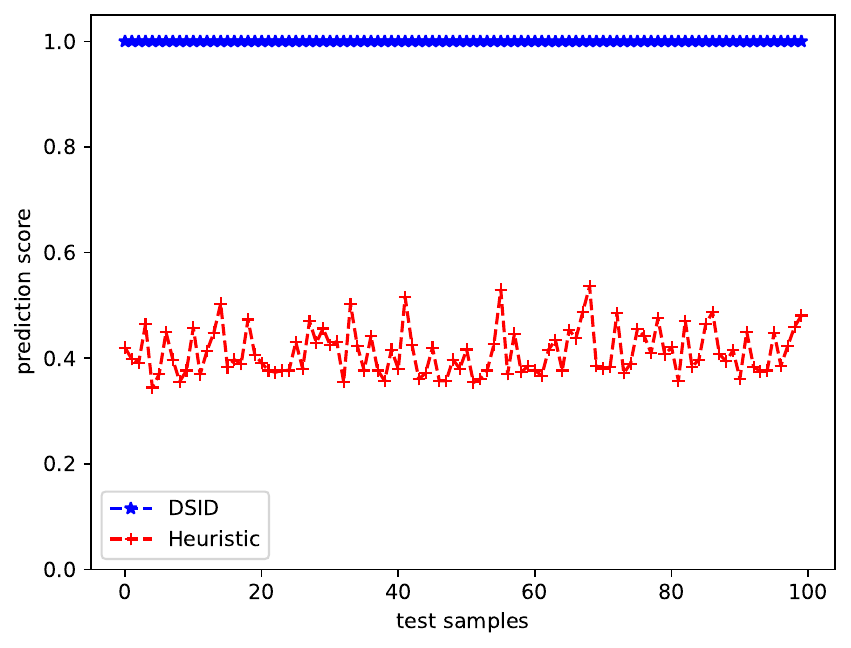}
    \caption{Evaluation of prediction score of DSID model and heuristic ML method using only 100 test samples.}\label{fig:pred1}
\end{figure}

\begin{figure}
    \centering
    \includegraphics[width=0.5\textwidth]{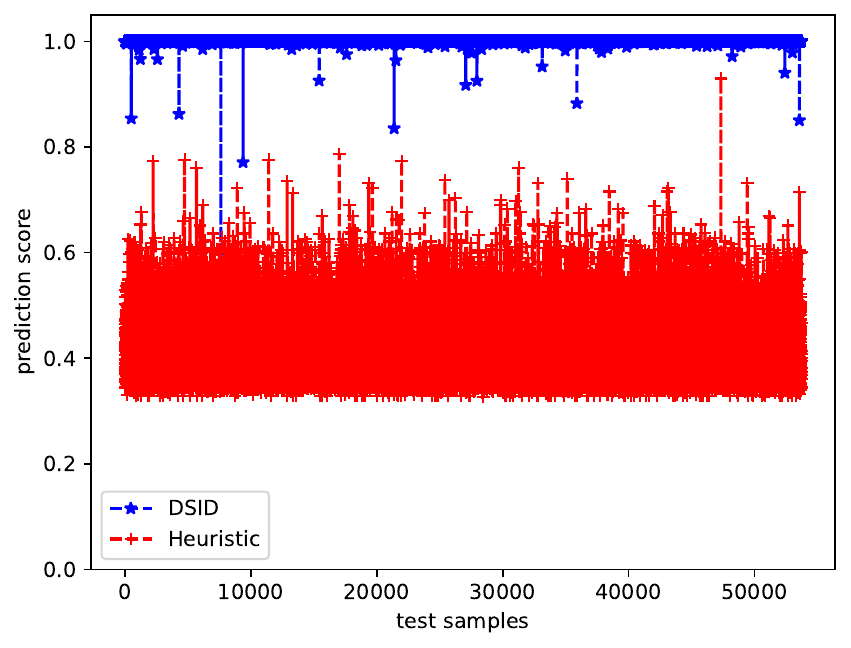}
    \caption{Comparing the prediction score of DSID model and heuristic ML method on all samples in the testing set.}\label{fig:pred2}
\end{figure}

\begin{table*}[htbp]
    \centering
        \caption{Probability distribution based on demographic symptom classification. For the class labels, indexing starts at C0 and ends at C26.}\label{tab:demographic_dist}
        \scriptsize
        \begin{tabular}{|c|c|c|c|c|c|c|c|c|c|c|c|c|c|c|} 
        \hline
          & C0 & C1 & C2 & C3 & C4 & C5 & C6 & C7 & C8 & C9 & C10 & C11 & C12 & C13\\ 
         \hline
         Female 0 - 9 Years & 0 & 0.03 & 0.01 & 0 & 0.03 & 0 & 0.01 & 0 & 0.03 & 0.02 & 0 & 0 & 0 & 0\\
         Female 10 - 19 Years & 0.04 & 0.06 & 0.07 & 0.03 & 0.02 & 0 & 0.07 & 0.01 & 0.05 & 0.04 & 0 & 0.02 & 0 & 0.01\\
         Female 20 - 29 Years & 0.16 & 0.11 & 0.16 & 0.12 & 0.07 & 0.01 & 0.18 & 0.03 & 0.13 & 0.09 & 0.02 & 0.05 & 0.05 & 0.02\\
         Female 30 - 39 Years & 0.12 & 0.07 & 0.1 & 0.1 & 0.06 & 0.01 & 0.12 & 0.06 & 0.08 & 0.07 & 0.06 & 0.03 & 0.09 & 0.03\\
         Female 40 - 49 Years & 0.13 & 0.06 & 0.09 & 0.11 & 0.06 & 0.04 & 0.11 & 0.06 & 0.07 & 0.08 & 0.08 & 0.04 & 0.13 & 0.02\\
         Female 50 - 59 Years & 0.13 & 0.07 & 0.09 & 0.11 & 0.08 & 0.07 & 0.11 & 0.11 & 0.08 & 0.11 & 0.1 & 0.05 & 0.16 & 0.06\\
         Female 60 - 69 Years & 0.08 & 0.05 & 0.05 & 0.07 & 0.07 & 0.08 & 0.06 & 0.11 & 0.07 & 0.09 & 0.12 & 0.12 & 0.16 & 0.12\\
         Female 70 - 79 Years & 0.04 & 0.04 & 0.02 & 0.04 & 0.06 & 0.11 & 0.03 & 0.06 & 0.05 & 0.07 & 0.06 & 0.06 & 0.05 & 0.09\\
         Female 80+ Years & 0.01 & 0.03 & 0 & 0.01 & 0.03 & 0.1 & 0 & 0.04 & 0.03 & 0.02 & 0.03 & 0.07 & 0.01 & 0.07\\
         Male 0 - 9 Years & 0 & 0.04 & 0.01 & 0.01 & 0.03 & 0 & 0.01 & 0 & 0.05 & 0.03 & 0 & 0.01 & 0 & 0\\
         Male 10 - 19 Years & 0.03 & 0.08 & 0.07 & 0.03 & 0.03 & 0 & 0.06 & 0 & 0.06 & 0.04 & 0 & 0.01 & 0.01 & 0.01\\
         Male 20 - 29 Years & 0.06 & 0.07 & 0.08 & 0.07 & 0.06 & 0.01 & 0.06 & 0.03 & 0.05 & 0.05 & 0.02 & 0.01 & 0.02 & 0.02\\
         Male 30 - 39 Years & 0.05 & 0.06 & 0.07 & 0.08 & 0.07 & 0.03 & 0.05 & 0.04 & 0.04 & 0.05 & 0.06 & 0.03 & 0.04 & 0.04\\
         Male 40 - 49 Years & 0.05 & 0.05 & 0.06 & 0.07 & 0.07 & 0.05 & 0.04 & 0.08 & 0.04 & 0.06 & 0.1 & 0.05 & 0.07 & 0.03\\
         Male 50 - 59 Years & 0.05 & 0.05 & 0.05 & 0.06 & 0.07 & 0.07 & 0.04 & 0.11 & 0.05 & 0.04 & 0.11 & 0.06 & 0.08 & 0.14\\
         Male 60 - 69 Years & 0.05 & 0.07 & 0.05 & 0.06 & 0.09 & 0.14 & 0.04 & 0.16 & 0.06 & 0.06 & 0.13 & 0.13 & 0.08 & 0.11\\
         Male 70 - 79 Years & 0.01 & 0.04 & 0.02 & 0.02 & 0.07 & 0.15 & 0.01 & 0.05 & 0.04 & 0.05 & 0.08 & 0.16 & 0.03 & 0.17\\
         Male 80+ Years & 0 & 0.02 & 0 & 0.01 & 0.03 & 0.13 & 0 & 0.05 & 0.02 & 0.02 & 0.03 & 0.11 & 0.01 & 0.07\\         
        \hline        
        \end{tabular}
        \medskip

        \scriptsize
        \begin{tabular}{|c|c|c|c|c|c|c|c|c|c|c|c|c|c|} 
        \hline
          & C14 & C15 & C16 & C17 & C18 & C19 & C20 & C21 & C22 & C23 & C24 & C25 & C26\\ 
         \hline
         Female 0 - 9 Years & 0 & 0 & 0.01 & 0 & 0 & 0 & 0 & 0 & 0 & 0 & 0 & 0 & 0\\
         Female 10 - 19 Years & 0 & 0 & 0 & 0.04 & 0.03 & 0.02 & 0.01 & 0 & 0 & 0.02 & 0.01 & 0.02 & 0\\
         Female 20 - 29 Years & 0 & 0.02 & 0.03 & 0.08 & 0.07 & 0.05 & 0.07 & 0 & 0.03 & 0.08 & 0.04 & 0.03 & 0.01\\
         Female 30 - 39 Years & 0.02 & 0.04 & 0.04 & 0.08 & 0.09 & 0.08 & 0.09 & 0.02 & 0.04 & 0.08 & 0.09 & 0.08 & 0.02\\
         Female 40 - 49 Years & 0.03 & 0.05 & 0.04 & 0.06 & 0.08 & 0.12 & 0.09 & 0.05 & 0.04 & 0.07 & 0.08 & 0.07 & 0.05\\
         Female 50 - 59 Years & 0.05 & 0.1 & 0.1 & 0.12 & 0.16 & 0.15 & 0.13 & 0.05 & 0.09 & 0.06 & 0.13 & 0.11 & 0.09\\
         Female 60 - 69 Years & 0.09 & 0.12 & 0.09 & 0.09 & 0.1 & 0.13 & 0.11 & 0.14 & 0.18 & 0.08 & 0.12 & 0.12 & 0.2\\
         Female 70 - 79 Years & 0.13 & 0.13 & 0.1 & 0.04 & 0.1 & 0.05 & 0.07 & 0.15 & 0.05 & 0.05 & 0.03 & 0.07 & 0.06\\
         Female 80+ Years & 0.1 & 0.05 & 0.07 & 0.02 & 0.01 & 0.02 & 0.02 & 0.06 & 0.07 & 0.07 & 0.03 & 0.02 & 0.03\\
         Male 0 - 9 Years & 0 & 0 & 0.01 & 0 & 0 & 0 & 0 & 0 & 0 & 0 & 0 & 0 & 0\\
         Male 10 - 19 Years & 0 & 0 & 0.01 & 0.01 & 0.01 & 0.01 & 0.01 & 0 & 0 & 0.03 & 0.02 & 0 & 0\\
         Male 20 - 29 Years & 0.01 & 0.02 & 0.02 & 0.04 & 0.03 & 0.03 & 0.01 & 0 & 0.03 & 0.03 & 0.02 & 0.02 & 0.01\\
         Male 30 - 39 Years & 0.01 & 0.04 & 0.04 & 0.06 & 0.05 & 0.05 & 0.03 & 0.04 & 0.01 & 0.06 & 0.07 & 0.07 & 0.03\\
         Male 40 - 49 Years & 0.04 & 0.07 & 0.04 & 0.07 & 0.06 & 0.08 & 0.06 & 0.05 & 0.03 & 0.04 & 0.06 & 0.05 & 0.07\\
         Male 50 - 59 Years & 0.09 & 0.05 & 0.07 & 0.07 & 0.06 & 0.08 & 0.06 & 0.06 & 0.09 & 0.07 & 0.07 & 0.1 & 0.1\\
         Male 60 - 69 Years & 0.13 & 0.13 & 0.12 & 0.12 & 0.09 & 0.09 & 0.13 & 0.14 & 0.08 & 0.05 & 0.13 & 0.12 & 0.12\\
         Male 70 - 79 Years & 0.14 & 0.11 & 0.11 & 0.06 & 0.02 & 0.03 & 0.06 & 0.16 & 0.13 & 0.11 & 0.08 & 0.11 & 0.16\\
         Male 80+ Years & 0.15 & 0.07 & 0.1 & 0.03 & 0.02 & 0.01 & 0.04 & 0.09 & 0.12 & 0.1 & 0.01 & 0.02 & 0.05\\         
        \hline        
        \end{tabular}  
        \medskip
\end{table*}    

\subsection{Symptoms Classification and Demographic Prediction}
The classes (or clusters) obtained from clustering the patients' symptoms using the Kmeans++ algorithm helps to reveal hidden details about the COVID-19 demographic distribution. We use Figure \ref{fig:cluster} to show the quality of clustering using the Kmeans++ algorithm, with each class being uniquely separated from other classes. The outcome of the clustering itself is not sufficient to extract useful and unknown information about the demographic distribution of the symptoms. Hence, we use the MLE parameter estimation method to compute the probability distribution of each demographic variable combination of age group (D) and gender (V), given each class label, for example:
$P(\text{age = 0 - 9 years}, \text{gender = Female} \mid \text{class = 1})$. 
Thus, each class maps to the probability distribution of the demographic variables given that class label. These results are stored locally, and can be retrieved using the class labels.

From Table \ref{tab:demographic_dist} which shows the demographic probability distribution conditioned on the class labels, we can observe that certain age groups and gender do not belong to some classes of symptom, as their probability is zero. This means that there is a demographic relationship to the way the symptoms are clustered. Our idea of using the class labels as signals to train the DSID model in a supervised learning manner is fruitful in its result as it helps us to learn the mapping relationship between patients' symptoms and the demographic distribution of their ground truth class. Essentially, we use the trained DSID model to predict the class of patients' symptoms and correspondingly obtain the demographic probability distribution. For the training of the DSID model, we used all the symptoms in Table \ref{tab:dataset} as input, that is, excluding the severity and demographic variables. We split the dataset into 0.8, 0.1 and 0.1 for training, validation and testing, respectively. The DSID model accuracy on the testing set is 99.99\% (as against the 41.15\% accuracy of a conventional heuristic algorithm) which further validates the quality of information obtained from the clustering stage. However, to obtain the age group and gender that a patient's symptom belongs to, we use:

\begin{equation}
   \theta^* = \underset{\theta}{\text{argmax}} \text{  }C_i (\theta), \quad i \in \{0, \cdots, 26\}
\end{equation}
where parameter $\theta$ denotes the probability scores (likelihood estimation) in the predicted class $C_i$ with $argmax$ referring to the age group and gender category with the highest probability score. For example, in Table \ref{tab:demographic_dist}, if a patient's symptoms class is predicted to be class 6 by the DSID model, then the gender and age group are highly probable to be Female and 20 -29 years, respectively.

There are other techniques to solve the demographic symptom classification problem. One popular method is to design a multi-output classification model in which the input features is mapped to multiple output categories; this is famous in the RNN (e.g., language translation, and texts from image) and CNN (e.g., image segmentation) domain. We show the strength of our method by comparing with a heuristic ML method which takes patients' symptoms as input, and outputs the predicted age class and predicted gender class, respectively. The heuristic model uses the same configuration in Table \ref{tab:param}, but with an exception at the output layer $\textbf{W}_6$. For the heuristic ML method, we trained two classifiers which share the same neural network architecture and weight parameters at the hidden layers, but have separate output layers and weights. To compute the validation accuracy using the heuristic ML, we average the accuracy scores of the two classifiers.

Figure \ref{fig:validation} shows the validation accuracy over the training epochs using our proposed DSID model and the heuristic ML method. The DSID model is quick to learn the mapping relationship between the input features and class labels, while the heuristic ML model cannot learn the mapping relationship over several training epochs. To show that our model does not overfit the data, we used out-of-sample (testing set) data to evaluate the performance, which is evident in Figure \ref{fig:pred1} and Figure \ref{fig:pred2} showing the prediction score on 100 test samples and the entire testing samples, respectively. The minimum and maximum prediction score for the DSID model are 0.6 and 1.0, respectively, while that of the heuristic ML model are 0.33 and 0.93, respectively. Note that the prediction score measures the confidence of the model in making predictions (this is not the same as model accuracy), and the interval for the prediction score is [0, 1].


\FloatBarrier
\section{Conclusion}
The proposed three-stage data-driven approach vis a vis Bayesian network structure learning, data clustering and supervised learning for COVID-19 severity explanation and demographic symptom classification, has shown tremendous benefits in this paper. We have clearly demystified the hidden truths about the causal relationships of the COVID-19 symptoms, and how they affect different demographics (i.e., age groups and gender). The approach adopted in this paper has not been used elsewhere to the best of the authors' knowledge, and thus opens up the research community to further explore probabilistic graphical models with machine learning to solve more challenging data science problems.

\section*{Declaration of Competing Interest}
The authors declare that they have no known competing financial interests or personal relationships that could have appeared to influence the work reported in this paper.

\section*{Acknowledgment}

This work is part of the Socially Responsible Modeling, Computation, and Design (SoReMo) research project in Illinois Institute of Technology, Chicago, founded by Prof. Sonja Petrovi\'c. 

\bibliographystyle{elsarticle-num}

\bibliography{article}

\end{document}